\documentclass[10pt,twocolumn,letterpaper]{article}

\usepackage[pagenumbers]{cvpr} 

\definecolor{cvprblue}{rgb}{0.21,0.49,0.74}
\usepackage[pagebackref,breaklinks,colorlinks,allcol
ors=cvprblue]{hyperref}
\usepackage{colortbl}
\usepackage{float}
\usepackage{multirow}
\usepackage{pifont}
\usepackage{amssymb}
\usepackage{calc}
\usepackage{xcolor}

\definecolor{LegendDarkGreen}{RGB}{0,128,0}
\definecolor{LegendGray}{RGB}{128,128,128}
\definecolor{LegendBrightGreen}{RGB}{0,255,0}
\definecolor{LegendBlue}{RGB}{0,0,255}
\definecolor{LegendDarkRed}{RGB}{128,0,0}

\def\paperID{322} 
\def\confName{CVPR}
\def\confYear{2026}

\title{UniChange:~Unifying~Change~Detection~with~Multimodal~Large~Language~Model}

\author{
Xu Zhang$^1$\thanks{Equal Contribution}\\
\and
Danyang Li$^{2*}$\\
\and
Xiaohang Dong$^1$\\
\and 
Tianhao Wu$^4$\\
\and
Hualong Yu$^{1}$\\
\and
Jianye Wang$^1$\\
\and
Qicheng Li$^1$\thanks{Corresponding Author}\\
\and
Xiang Li$^{2,3}$\\
\and
$^1$TMCC, Computer Science, Nankai University ~~~~ $^2$VCIP, Computer Science, Nankai University ~~~~ \\
$^3$NKIARI, Futian, Shenzhen ~~~~ $^4$CMEE, Sichuan Agricultural University\\
\textit{\{xu\_zhang, danyang.li\}@mail.nankai.edu.cn, liqicheng@nankai.edu.cn}}

\begin{document}
\maketitle
\begin{abstract}
Change detection (CD) is a fundamental task for monitoring and analysing land cover dynamics. While recent high performance models and high quality datasets have significantly advanced the field, a critical limitation persists. Current models typically acquire limited knowledge from single-type annotated data and cannot concurrently leverage diverse binary change detection (BCD) and semantic change detection (SCD) datasets. This constraint leads to poor generalisation and limited versatility. The recent advancements in Multimodal Large Language Models (MLLMs) introduce new possibilities for a unified CD framework. We leverage the language priors and unification capabilities of MLLMs to develop UniChange, the first MLLM-based unified change detection model. UniChange integrates generative language abilities with specialised CD functionalities. We introduce three special tokens: [T1], [T2], and [CHANGE], utilising their embeddings as the key to query variations. This approach successfully accommodates both BCD and SCD tasks. Furthermore, UniChange utilises text prompts to guide the identification of change categories, eliminating the reliance on predefined classification heads. This design allows UniChange to effectively acquire knowledge from multi-source datasets, even when their class definitions conflict. Experiments on four public benchmarks (WHU-CD, S2Looking, LEVIR-CD+, and SECOND) demonstrate SOTA performance, achieving IoU scores of 90.41, 53.04, 78.87, and 57.62, respectively, surpassing all previous methods. The code is available at \url{https://github.com/Erxucomeon/UniChange}.
\end{abstract}    
\section{Introduction}
\label{sec:intro}

Change detection (CD) is the process of observing and analysing multi-temporal remote sensing images to identify changes in land surface cover. It serves as a fundamental task in the remote sensing field and a cornerstone of modern geospatial analysis. Change detection plays a vital role in numerous applications, ranging from sustainable urban planning \cite{wellmann2020remote} and natural disaster assessment \cite{zheng2021building} to ecological monitoring \cite{singh1986change} and land resource management \cite{asadzadeh2022uav}. Depending on the required level of detail, CD tasks are typically categorised into two primary sub-domains: binary change detection (BCD) and semantic change detection (SCD). BCD, the simpler task, primarily identifies the locations where change has occurred, but may focus on a single object of interest or not specify the nature of these transformations. While useful for automated monitoring, it provides relatively coarse-grained information. In contrast, SCD is a more complex and refined task. It not only determines the location of change but also identifies the ``from-to'' semantic transition \cite{mei2024scd} (e.g., from ``forest'' to ``urban land''). This granular understanding of semantic shifts offers richer insights, providing more robust decision support for the aforementioned applications.

\begin{figure}[t]
  \centering
   \includegraphics[width=0.95\linewidth]{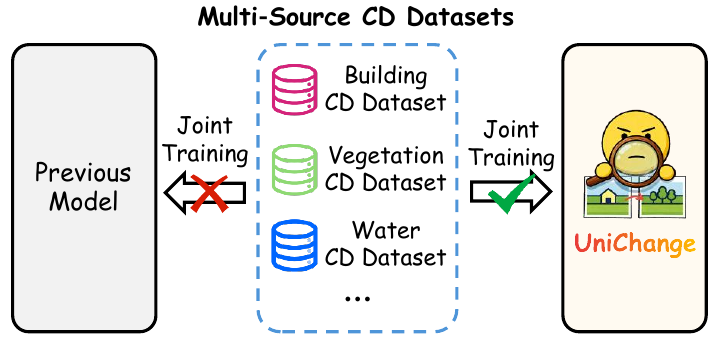}
   \caption{Inconsistency~in~multi-source~change~detection~datasets.}
   \label{fig:Inconsistency datasets}
\end{figure}

In the deep learning era, methods such as FC-Siam-diff \cite{daudt2018fully} and IFN \cite{zhang2020deeply} establish the siamese network architecture as the dominant paradigm for BCD. Subsequent research shifts focus toward enhancing the fusion and interaction of dual-temporal features. 
BiT \cite{chen2022remote} fuses dual-temporal features using token-space context modeling and feature refinement.
Concurrently, Transformer-based architectures \cite{bandara2022transformer, zhang2022swinsunet, xu2024hybrid} are proposed to fuse and model long-range dependencies between these features at various stages within the encoder. Changer \cite{fang2023changer} pioneers a dual-temporal feature exchange paradigm by inserting a series of feature interaction layers directly within the feature extractor. Furthermore, recent works like SAM-CD \cite{ding2024adapting} have begun to adapt Vision Foundation Models (VFMs) \cite{kirillov2023segment, zhao2023fast, ravi2024sam} for the BCD task. However, a fundamental limitation of these methods is the necessity of training a specialised model for each individual dataset. RSBuilding \cite{wang2024rsbuilding} partially mitigates this issue by leveraging a VFM, enabling joint training on multiple building datasets for both building extraction and change detection. However, its framework is essentially tailored to building-related tasks and cannot be extended to other categories of land cover change.

Compared to BCD, SCD is considered a more challenging task. Its core complexity lies in the fact that SCD must not only localise changes but also identify specific ``from-to'' semantic transitions between land cover categories. To address this challenge, the early HRSCD \cite{daudt2019multitask} adopts a decoupling strategy, jointly training semantic segmentation and binary change detection, which is subsequently refined by MTSCD-Net \cite{cui2023mtscd}. Subsequent studies have turned to deeper feature interaction and temporal modelling. Bi-SRNet \cite{ding2022bi} improves semantic change detection via dual-temporal reasoning and consistency modelling. Furthermore, SCD-SAM \cite{mei2024scd} leverages the representational capabilities of VFMs and designs a refined dual-encoder and dual-decoder architecture. The SCD task's inherent specificity and complexity necessitate a specialised architecture, thereby hindering architectural unification between BCD and SCD tasks.

\begin{figure}[t]
  \centering
   \includegraphics[width=0.88\linewidth]{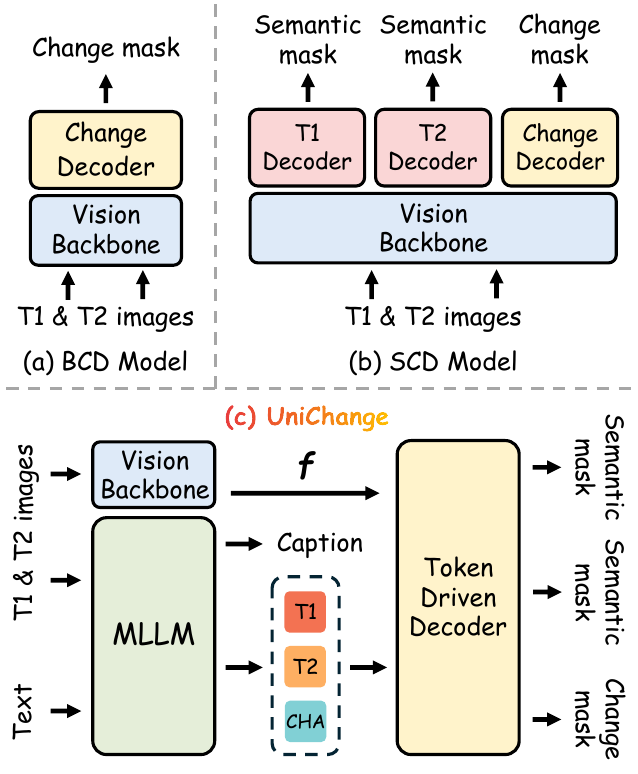}
   \caption{Inconsistency in model architecture.}
   \label{fig:Inconsistency architecture}
\end{figure}

By analysing the previous BCD and SCD methods, we have identified two problems that have persisted in the field. The first is dataset incompatibility. As shown in \cref{fig:Inconsistency datasets}, a positive sample in one dataset (e.g., ``building'') may be explicitly defined as a negative background sample in another dataset (e.g., ``vegetation''). This semantic conflict prevents traditional models from being jointly trained on such multi-source datasets. Secondly, the different requirements of BCD and SCD lead to incompatible model architectures, as shown in \cref{fig:Inconsistency architecture}(a) and \cref{fig:Inconsistency architecture}(b). This has led to a proliferation of highly specialised models, each with limited knowledge from a single annotation type of dataset (BCD or SCD). This mutually independent design approach results in poor model generalisation and severely limits the versatility of the CD system in practical applications.

To address the fundamental challenges within the change detection domain, we propose UniChange. UniChange is a unified change detection model based on the multimodal large language model. Not only does UniChange simultaneously support both BCD and SCD tasks, but it also enables joint training using multi-source datasets. The architecture of UniChange is shown in \cref{fig:Inconsistency architecture}(c). We introduce three special tokens [T1], [T2], and [CHANGE] to enable the model to handle both BCD and SCD tasks concurrently. Furthermore, UniChange leverages textual cues to guide change region classification, thereby eliminating the reliance on predefined classification heads found in traditional approaches. This design facilitates the extraction of unified knowledge from diverse, multi-source change detection datasets, even when semantic category conflicts exist between datasets. Overall, our contributions are as follows:

\begin{itemize}
\item[$\bullet$] We propose UniChange, the first unified framework for change detection based on the MLLM. It is designed to simultaneously accommodate both binary change detection (BCD) and semantic change detection (SCD) tasks within a single end-to-end model, whilst enabling training using multi-source change detection datasets.
\item[$\bullet$] We introduce a Token Driven Decoder strategy. The strategy incorporates three special tokens into MLLM: $\text{[T1]}$, $\text{[T2]}$ and $\text{[CHANGE]}$. It enables the model to be trained simultaneously on change detection tasks and the semantic understanding of dual-temporal images.
\item[$\bullet$] Comprehensive experiments demonstrate that UniChange achieves state-of-the-art performance, significantly outperforming previous methods on both the BCD and SCD benchmark datasets.
\end{itemize}
\section{Related Work}
\subsection{Change Detection}

The field of change detection (CD) is broadly divided into two sub-fields: binary change detection (BCD) and the more complex semantic change detection (SCD). The latter is more challenging as it requires identifying the ``from-to'' semantic information, not just the presence of change.

Early deep learning methods for BCD are dominated by CNN-based architectures. FC-Siam-diff \cite{daudt2018fully} is one of the early methods to utilise the paradigm of comparing features from two weight-sharing encoders. IFN \cite{zhang2020deeply} extends this by adding deep supervision and attention mechanisms. To better model global context, Transformer-based methods are introduced, including BiT \cite{chen2022remote}, ChangeFormer \cite{bandara2022transformer}, SwinSUNet \cite{zhang2022swinsunet} and Changer \cite{fang2023changer}. Other notable works, such as DMINet \cite{feng2023change}, and HATNet \cite{xu2024hybrid}, further explore hybrid architectures and deep feature interaction. More recently, VFM-based methods have emerged. SAM-CD \cite{ding2024adapting} focuses on adapting VFMs \cite{kirillov2023segment, zhao2023fast, ravi2024sam} to the BCD task, while RSBuilding \cite{wang2024rsbuilding} leverages VFMs to unify building-specific extraction and change detection tasks. However, these methods are difficult to jointly train on diverse and multi-class change detection datasets, limiting their generalisation.

Compared with BCD task, SCD task is more complex. HRSCD \cite{daudt2019multitask} and MTSCD-Net \cite{cui2023mtscd} decouple the task into binary change detection and semantic segmentation subtasks, integrating them within a multi-task framework for refined feature association. SCDNet \cite{peng2021scdnet} employs siamese networks and further utilises attention mechanisms to refine multi-scale difference features, while Bi-SRNet \cite{ding2022bi} proposes cross-temporal semantic reasoning blocks to enhance feature consistency. Most recently, VFM-based adaptation emerges. SCD-SAM \cite{mei2024scd} adapts VFMs by designing a specialised dual-encoder and dual-decoder system specifically for the SCD task. The differences between BCD and SCD tasks lead to inconsistency between their model architectures, resulting in poor versatility.

\subsection{Multimodal Large Language Model}
Initial MLLMs, such as LLaVA \cite{liu2023visual}, MiniGPT-4 \cite{zhu2023minigpt}, and InstructBLIP \cite{dai2023instructblIP}, focus on tasks such as image captioning and visual question answering (VQA). Subsequent research has further developed this capability, enabling it to support refined visual understanding, perception and generation at both regional and pixel levels \cite{zhang2024gpt4roi, you2023ferret, peng2024grounding, lai2024lisa, li2025visual, wu2025representation}. While transformative, their application to remote sensing (RS) has been challenging. RS-specific MLLMs such as RSGPT \cite{hu2025rsgpt}, GeoChat \cite{kuckreja2024geochat}, and GeoPixel \cite{shabbir2025geopixel} have emerged, but these methods largely focus on single-image interpretation. This architecture, which focuses on single images, makes it ill-suited for comparative, dual-temporal analysis. It leaves a significant gap in the ability to perform pixel-level grounding for change detection.

\begin{figure*}[t]
  \centering
   \includegraphics[width=1\linewidth]{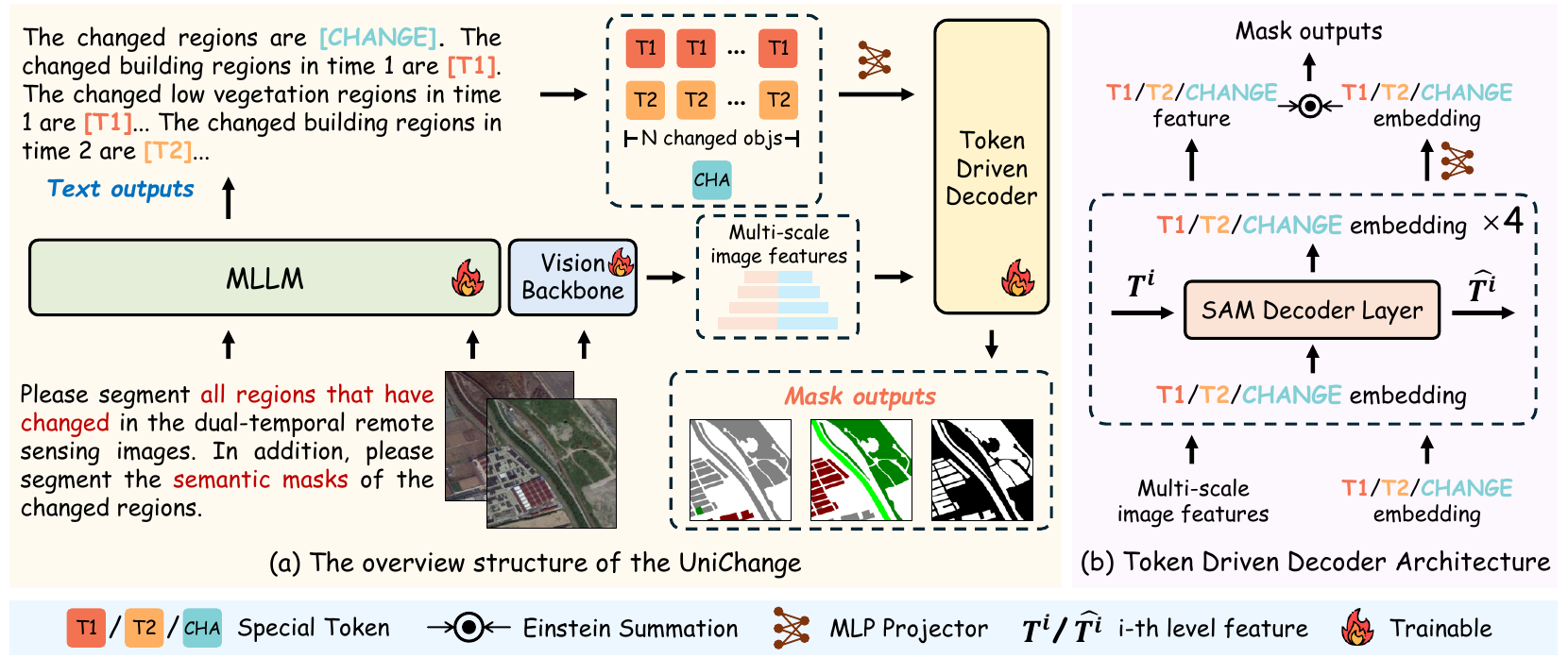}
   \caption{The overview structure of UniChange is shown in (a). UniChange generates text outputs based on text instructions and dual-temporal remote sensing images. The text outputs incorporate a series of special tokens ([T1], [T2], and [CHANGE]). Concurrently, the dual-temporal remote sensing images undergo feature extraction via the Vision Backbone. These features, alongside the embeddings corresponding to the special tokens, are fed into the Token Driven Decoder. Ultimately, UniChange generates the masks corresponding to the text instructions. The overall structure of the Token Driven Decoder is shown in (b). It receives dual-temporal remote sensing image features and special token embeddings, generating corresponding masks.}
   \label{fig:main}
\end{figure*}

\section{Method}
In this section, we first define the unified change detection task. As detailed in \cref{Task Definition}, this task involves dual-temporal remote sensing images and textual queries. Subsequently, we present the UniChange architecture in detail in \cref{Architecture}. This architecture comprises an MLLM for understanding changes, a Vision Backbone for extracting features from dual-temporal remote sensing images, and a specialised Token Driven Decoder. Finally, we detail the training strategy in \cref{Training Strategy}, including the unified loss functions designed to manage both binary and semantic constraints.

\subsection{Task Definition}
\label{Task Definition}
We define the task of unified change detection as follows. Let the input consist of a pair of dual-temporal remote sensing images, $x_{img1}, x_{img2} \in R^{H \times W \times C}$, captured at times $T_1$ and $T_2$, where $H, W,$ and $C$ represent the height, width, and channel dimensions, respectively. An accompanying text input, $x_{txt}$, provides the change query instruction. This instruction may be a binary change query (e.g., ``Please segment all areas that have undergone change.'') or a more complex semantic change query (e.g., ``Please segment the semantic masks of the changed areas.''). The core goal is to generate a set of masks ($\hat{M}$) that accurately adhere to the semantic instructions provided by the query $x_{txt}$. This overall process can be formulated as
\begin{equation}
\hat{M} = \Phi_{UniChange}(x_{img1}, x_{img2}, x_{txt}),
\end{equation}
where $\Phi_{UniChange}$ represents the entire unified change detection model.

\subsection{Architecture}
\label{Architecture}
\textbf{Embedding as Change.} 
Current change detection (CD) models suffer from two fundamental problems: dataset incompatibility due to semantic conflicts across different sources, and architectural divergence between specialised binary change detection (BCD) and semantic change detection (SCD) tasks. This fragmentation prevents joint training, leading to poor generalisation and limited versatility. Therefore, we propose UniChange. This framework implements our proposed ``embedding as change'' paradigm. This paradigm integrates generative language abilities with specialised CD functionalities, thereby achieving a true unification of multi-source datasets and multi-tasks. The framework of our method is illustrated in \cref{fig:main}(a). MLLM takes the dual-temporal remote sensing images $x_{img1}$, $x_{img2}$ and a given textual instruction $x_{txt}$ as input. To unify both binary and semantic change detection within a single framework, we introduce three specialised tokens into the MLLM's vocabulary: [T1], [T2] and [CHANGE]. Conditioned on the textual instruction $x_{txt}$, the MLLM autoregressively generates the response sequence $y_{txt}$. These tokens are strategically placed within this sequence in accordance with the specific task query. This process can be formulated as
\begin{equation}
y_{txt} = \mathcal{F}_{MLLM}(x_{img1}, x_{img2}, x_{txt}).
\end{equation}

When the MLLM intends to generate masks for a specific change (e.g., the binary change mask or a semantic mask for $T_1$ and $T_2$), its output $y_{txt}$ includes the corresponding special token (e.g., [T1], [T2] or [CHANGE]). Subsequently, we extract the MLLM last-layer embedding $h_{task}$ corresponding to the position of a specific token ([T1], [T2] or [CHANGE]) within $y_{txt}$. This raw embedding is then projected by a dedicated $MLP(\cdot)$ layer to make it compatible with the downstream vision module's feature space:
\begin{equation}
\label{eq:task}
\hat{h}_{task} = MLP(h_{task}).
\end{equation}

These projected sparse embeddings  (where each is individually generated as $\hat{h}_{task}$ via \cref{eq:task}) serve as the dynamic, instruction-guided queries. 

Concurrently, the vision backbone $\mathcal{F}_{enc}$ first extracts base dense visual features from the dual-temporal remote sensing images. A feature pyramid is then constructed using max-pooling and transposed convolutional layers. It generates four levels of multi-scale image features for each temporal image. This results in two multi-scale image feature sets, denoted as $\{F_1^i\}$ and $\{F_2^i\}$, where $i \in \{1, 2, 3, 4\}$. Finally, these dual-temporal visual features, along with the projected task embeddings ${\hat{h}}_{task}$, are fed into the token driven decoder $\mathcal{F}_{dec}$ to generate the final mask $\hat{M}_{task}$. The entire process can be described as
\begin{equation}
\begin{aligned}
    \{F_1^i\}, \{F_2^i\} = \mathcal{F}_{enc}(x_{img1}, x_{img2}), \\
    \hat{M}_{task} = \mathcal{F}_{dec}(\{F_1^i\}, \{F_2^i\}, {\hat{h}}_{task}).
\end{aligned}
\end{equation}

We employ a hybrid fine-tuning strategy. Only the language decoder of the MLLM undergoes LoRA fine-tuning \cite{hu2022lora}. All other components of the model are fully parameterised for fine-tuning.

\textbf{Token Driven Decoder.} 
Inspired by RSBuilding, the token driven decoder, $\mathcal{F}_{dec}$, is specifically architected to translate the MLLM's high-level instructions into pixel-level segmentation masks for the change detection task. Its architecture is illustrated in \cref{fig:main}(b). This decoder is employed to process the dual-temporal remote sensing image features and the projected special token embeddings. Specifically, the projected embeddings $\{\hat{h}_{t1}, \hat{h}_{t2}, \hat{h}_{change}\}$ (obtained from \cref{eq:task}) serve as the initial task queries. Let $E^0$ represent the concatenation ($\Phi_{Cat}$) of these initial queries:
\begin{equation}
 E^0 = \Phi_{Cat}(\hat{h}_{t1}, \hat{h}_{t2}, \hat{h}_{change}),
\end{equation}
the concatenated task query $E^0$ is then hierarchically refined through four decoder layers. Each decoder layer processes its corresponding multi-scale image features $F_1^i$ and $F_2^i$ by flattening ($\Phi_{Flat}$) them into token sequence, and concatenating ($\Phi_{Cat}$) them into a unified visual sequence $T^i$:
\begin{align}
T^i = \Phi_{Cat}(\Phi_{Flat}(F_1^i), \Phi_{Flat}(F_2^i)).
\end{align}

The SAM Decoder Layer then refines both the task query and the visual sequence. The task query $E^{i-1}$ (from the previous level) first attends to itself ($\mathcal{A}_{self}$), then attends to the visual sequence $T^i$ ($\mathcal{A}_{cross}$), and is subsequently processed by an FFN ($\mathcal{F}_{FFN}$). Next, the visual sequence $T^i$ is also updated by attending to this newly refined task query ($\mathcal{A}_{cross}$). The positional encodings are added to queries and keys in all attention operations. This process is defined as
\begin{equation}
\begin{gathered}
    E^{i} = \mathcal{F}_{FFN}(\mathcal{A}_{cross}(\mathcal{A}_{self}(E^{i-1}), T^i)), \\
    \hat{T}^{i} = \mathcal{A}_{cross}(T^i, E^{i}).
\end{gathered}
\end{equation}

The refined task query $E^{4}$ is taken from the final level. The query and the refined visual sequences $\{\hat{T}^{i}\}$ from all levels are then used for the final mask generation.

The visual sequences $\{\hat{T}^{i}\}$ from each level are split and reshaped ($\Phi_{Split/Reshape}$) back into their corresponding 2D dual-temporal feature maps, $\{\hat{F}_1^i\}$ and $\{\hat{F}_2^i\}$: 
\begin{gather}
    \{\hat{F}_1^i\}, \{\hat{F}_2^i\} = \Phi_{Split/Reshape}(\{\hat{T}^i\}) \label{eq:feat_process}.
\end{gather}

Next, three feature sets are prepared: $\{\hat{F}_1^i\}$, $\{\hat{F}_2^i\}$, and their element-wise difference $\{\hat{F}_1^i - \hat{F}_2^i\}$. All three sets are then uniformly processed by a three-step procedure: upsampling ($\Phi_{Up}$), concatenation ($\Phi_{Cat}$), and fusion ($\Phi_{fuse}$). This derives $F_{t1}$, $F_{t2}$, and $F_{change}$, respectively:
\begin{equation}
\begin{gathered}
    F_{t1} = \Phi_{fuse}(\Phi_{Cat}(\{\Phi_{Up}(\hat{F}_1^i)\})), \\
    F_{t2} = \Phi_{fuse}(\Phi_{Cat}(\{\Phi_{Up}(\hat{F}_2^i)\})), \\
    F_{change} = \Phi_{fuse}(\Phi_{Cat}(\{\Phi_{Up}(\hat{F}_1^i - \hat{F}_2^i)\})). \label{eq:feat_sub}
\end{gathered}
\end{equation}

Concurrently, the refined task query $E^4$ from the final decoder level is split and projected ($\Phi_{Split/Proj}$) into the individual task embeddings $\hat{e}_{t1}$, $\hat{e}_{t2}$, and $\hat{e}_{change}$:
\begin{equation}
\begin{aligned}
    \hat{e}_{t1}, \hat{e}_{t2}, \hat{e}_{change} &= \Phi_{Split/Proj}(E^4).
\end{aligned}
\end{equation}

Finally, these projected embeddings are used to filter their corresponding features ($F_{t1}$, $F_{t2}$, $F_{change}$) to generate the final masks:
\begin{equation}
\begin{aligned}
    \hat{M}_{task} &= \mathcal{M}_{gen}(F_{task}, \hat{e}_{task}), \\
    & \quad \text{task} \in \{t1, t2, change\}
\end{aligned}
\end{equation}
where $\mathcal{M}_{\text{gen}}$ represents the mask generation function. This function performs Einstein Summation.

This unified structure allows UniChange to flexibly generate the corresponding masks based on user instructions.

\subsection{Training Strategy}
\label{Training Strategy}

Our method is trained end-to-end to jointly optimise UniChange. The overall objective function, $\mathcal{L}_{total}$, which guides the entire optimisation process, is defined as a summation of all contributing loss terms:
\begin{equation}
    \mathcal{L}_{total} = \mathcal{L}_{txt} + \mathcal{L}_{mask}.
    \label{eq:12}
\end{equation}

In \cref{eq:12}, $\mathcal{L}_{txt}$ is the standard autoregressive Cross-Entropy loss applied to the MLMM's generated token sequence, optimising its next-token prediction capability. 

The comprehensive mask loss, $\mathcal{L}_{mask}$, is designed to enforce pixel-level accuracy across the unified change detection task. This loss is defined as the summation of four distinct components:
\begin{equation}
    \resizebox{.88\hsize}{!}{$
    \mathcal{L}_{mask} = \lambda_{BCE}\mathcal{L}_{BCE} + \lambda_{Dice}\mathcal{L}_{Dice} + \lambda_{SS}\mathcal{L}_{SS} + \lambda_{SC}\mathcal{L}_{SC}.
    $}
    \label{eq:13}
\end{equation}

In \cref{eq:13}, $\lambda_{BCE}$, $\lambda_{Dice}$, $\lambda_{SS}$, and $\lambda_{SC}$ serve as the weighting coefficients for the Binary Cross-Entropy loss, Dice loss, Semantic Segmentation loss, and Semantic Change loss, respectively. $\mathcal{L}_{BCE}$ and $\mathcal{L}_{Dice}$ are the Binary Cross-Entropy loss and the Dice loss, respectively. Both of them optimise the model's prediction by applying supervision between the predicted change masks and the ground truth change masks. $\mathcal{L}_{SS}$ is the Cross-Entropy loss for Semantic Segmentation, which computes pixel-wise classification errors across all spatial locations between the dual-temporal predictions and their respective ground truth labels, thereby penalizing misclassification across all semantic categories. $\mathcal{L}_{SC}$ is the Semantic Change loss, designed to enhance the discriminability of dual-temporal feature representations. This loss calculates a cosine embedding distance between the semantic feature maps based on the binary ground truth masks. Specifically, it enforces similarity for unchanged regions and encourages divergence for changed regions.

Crucially, during training on BCD datasets, the semantic constraint losses, $\mathcal{L}_{SS}$ and $\mathcal{L}_{SC}$, are set to zero. They are only computed when training on SCD datasets. By optimising this comprehensive objective function, UniChange ensures both robust and effective alignment between textual instructions and pixel-level perception, simultaneously achieving high accuracy in both binary and multi-class semantic change detection tasks.
\section{Experiments}
\subsection{Experimental Setting}
We employ LLaVA-7B-v1-1 \cite{liu2023visual} as the base MLMM. For Vision Backbone, we adopt the RSBuilding-ViT-L \cite{wang2024rsbuilding} backbone, which is a SAM backbone structure pre-trained on remote sensing imagery.

UniChange is trained for 10 epochs (with 400 steps per epoch) using 4 NVIDIA 80G H100 GPUs. We employ the AdamW \cite{loshchilov2017decoupled} optimizer with a base learning rate of $5 \times 10^{-5}$. The training is performed using a per-device batch size of 1 and a gradient accumulation step of 8, leveraging the deepspeed \cite{rasley2020deepspeed} engine for efficiency.

The comprehensive loss function, which is introduced in \cref{Training Strategy}, utilises several weighting coefficients to balance its multiple objective components. Specifically, the mask loss components are weighted with $\lambda_{BCE}$ set to $2.0$, $\lambda_{Dice}$ set to $0.5$, $\lambda_{SS}$ set to $0.5$, and $\lambda_{SC}$ set to 1.0.

\begin{table*}[t]
\small
\setlength{\tabcolsep}{7pt}
\renewcommand\arraystretch{1.05}
\centering
\caption{Comparison with existing SOTA methods on WHU-CD, S2Looking and LEVIR-CD+.}
\label{tab:bcd}
\begin{tabular}{c|cccc|cccc|cccc}
\toprule
\multirow{2}{*}{Methods} & \multicolumn{4}{c|}{WHU-CD} & \multicolumn{4}{c|}{S2Looking} & \multicolumn{4}{c}{LEVIR-CD+} \\
\cmidrule(l){2-13}
\multicolumn{1}{c|}{} & P & R & F1 & IoU & P & R & F1 & IoU & P & R & F1 & IoU \\
\midrule
FC-Siam-diff \cite{daudt2018fully}  & 48.84         & 88.96 & 63.06 & 46.05 & \textbf{83.49} & 32.32 & 46.60 & 30.38 & 80.88 & 77.65 & 79.23 & 65.60 \\
DASNet \cite{chen2020dasnet}        & 83.77         & 91.02 & 87.24 & 77.37 & 45.06          & 48.71 & 47.29 & 30.97 & 77.51 & 78.03 & 77.77 & 63.63 \\
SNUNet \cite{fang2021snunet}        & 91.28         & 87.25 & 89.22 & 80.54 & 45.25          & 50.60 & 47.78 & 31.39 & 78.90 & 78.23 & 78.56 & 64.69 \\
BIT \cite{chen2022remote}           & 91.56         & 87.84 & 89.66 & 81.26 & 70.26          & 56.53 & 62.65 & 45.61 & 82.37 & 79.73 & 81.03 & 68.11 \\
ChangeFormer \cite{bandara2022transformer} & 91.76         & 84.85 & 88.17 & 78.85 & 70.73          & 49.25 & 58.07 & 40.91 & 84.29 & 80.81 & 82.51 & 70.23 \\
RFL-CDNet \cite{gan2024rfl}     & 91.33         & 91.46 & 91.39 & 84.15 & 65.72          & 60.82 & 63.17 & 46.17 & 79.95 & 84.04 & 81.94 & 69.41 \\
SAM-CD \cite{ding2024adapting}       & \textbf{96.87} & 85.67 & 90.92 & 83.35 & 72.80          & 58.92 & 65.13 & 48.29 & 79.71 & 81.35 & 81.96 & 69.43 \\
Meta-CD \cite{gao2025combining}      & 89.00         & 90.35 & 89.67 & 81.27 & 74.08          & 54.00 & 62.47 & 45.42 & 80.17 & 84.13 & 82.10 & 69.93 \\
SA-CDNet \cite{gan2025detect}       & 95.29         & 93.67 & 94.47 & 89.52 & 81.28          & 56.24 & 66.48 & 49.79 & 85.55 & 83.44 & 84.43 & 73.06 \\
ChangeCLIP \cite{dong2024changeclip} & 96.02 & 93.58 & 94.78 & 90.08 & - & - & - & - & \textbf{88.46} & 83.90 & 86.12 & 75.63 \\
Changer \cite{fang2023changer}            & -         & - & - & - & 73.01          & 62.04 & 67.08 & 50.47 & - & - & - & - \\
LSKNet \cite{li2025lsknet}            & -         & - & - & - & 71.90          & 63.64 & 67.52 & 50.96 & - & - & - & - \\
SFCD-Net \cite{zhang2024integrating}    & 95.42         & 93.14 & 94.27 & 89.16 & 72.60          & 61.70 & 66.71 & 50.05 & 87.25 & 85.65 & 86.44 & 76.12 \\
TTP \cite{chen2024time}             & 96.05         & 92.76 & 94.37 & 89.34 & 73.51          & 62.19 & 67.38 & 50.80 & 85.81 & 84.36 & 85.08 & 74.03 \\
\rowcolor{blue!5}
\textbf{UniChange} &
  95.83 &
  \textbf{94.11} &
  \textbf{94.96} &
  \textbf{90.41} &
  73.48 &
  \textbf{65.60} &
  \textbf{69.32} &
  \textbf{53.04} &
  87.88 &
  \textbf{88.49} &
  \textbf{88.19} &
  \textbf{78.87} \\ \bottomrule
\end{tabular}
\end{table*}

\subsection{Datasets}
\textbf{BCD Datasets.} We evaluate the binary change detection (BCD) capability of UniChange on three widely adopted remote sensing datasets: WHU-CD \cite{ji2019fully}, S2Looking \cite{shen2021s2looking}, and LEVIR-CD+ \cite{chen2020spatial}. The WHU-CD dataset consists of a pair of large dual-temporal remote sensing images, each with a size of 32507 $\times$ 15354. For standardised evaluation, we crop these large images into 1024 $\times$ 1024 patches. We then split these patches into training, validation, and testing sets following an 8:1:1 ratio. The S2Looking dataset provides a total of 5000 image pairs captured from side-looking rural areas globally. These image pairs are divided into 3500 for training, 1000 for testing and 500 for validation. Finally, LEVIR-CD+ is an extended version of LEVIR-CD. It provides 985 pairs of 1024 $\times$ 1024 images, primarily focusing on building changes. We utilise the original 1024 $\times$ 1024 image size for training and testing, which provides 637 training patches and 348 testing patches. These datasets collectively allow for robust assessment across different resolutions and geographic regions.

\textbf{SCD Datasets.} We evaluate the semantic change detection (SCD) capability of UniChange on the SECOND dataset \cite{yang2022asymmetric}. The dataset is specifically designed for fine-grained change analysis, requiring models to simultaneously localise changes and classify semantic transitions. The dataset comprises 4662 pairs of dual-temporal remote sensing images with a 512 $\times$ 512 resolution. It covers six detailed land-cover categories: building, low vegetation, tree, water, playground, and bare ground. The predominance of unchanged pixels in the dataset creates a significant class imbalance, which places higher demands on the model's generalisation capability.

\subsection{Evaluation Metrics}
\textbf{BCD Metrics.} To evaluate $\text{BCD}$ performance, we employ four standard pixel-level metrics: Precision (P), Recall (R), F1 score (F1), and Intersection over Union ($\text{IoU}$). Precision measures the ratio of correctly predicted change pixels to all predicted change pixels. Recall measures the ratio of true positive pixels to all ground-truth positive pixels. The $\text{F1}$ score is the harmonic mean of $\text{P}$ and $\text{R}$. $\text{IoU}$ quantifies the overlap between predicted and ground-truth change regions.

\textbf{SCD Metrics.} To evaluate $\text{SCD}$ performance, we employ three metrics: $\text{mIoU}$, $\text{F}_{scd}$, and $\text{SeK}$. $\text{mIoU}$ evaluates the overall segmentation quality. $\text{F}_{scd}$ quantifies semantic transition accuracy using the harmonic mean of precision and recall. $\text{SeK}$ measures semantic discrimination, mitigating the impact of the unchanged class. UniChange's SCD performance is comprehensively evaluated using \ensuremath{\text{BCD}} metrics (\ensuremath{\text{IoU}}, \ensuremath{\text{F1}} (\ensuremath{\text{F}_{bcd}})) and \ensuremath{\text{SCD}} metrics (\ensuremath{\text{mIoU}}, \ensuremath{\text{F}_{scd}}, \ensuremath{\text{SeK}}). Further details on the $\text{BCD}$ and $\text{SCD}$ metrics are provided in \cref{appendix:eva_met} of the supplementary materials.

\subsection{Results}
\textbf{BCD Results.} As shown in \cref{tab:bcd}, UniChange achieves the best F1 score and IoU score across all three evaluated benchmarks: WHU-CD, S2Looking, and LEVIR-CD+. On the WHU-CD dataset, our method achieves top $\text{F1}$ and $\text{IoU}$ scores of 94.96 and 90.41. On the S2Looking dataset, our method obtains an F1 score of 69.32 and an IoU score of 53.04, exceeding the LSKNet result (67.52 F1 score, 50.96 IoU score). Furthermore, on the LEVIR-CD+ dataset, our approach demonstrates superior performance, setting a new SOTA F1 score of 88.19 and IoU score of 78.87, outperforming the second-best model, SFCD-Net (86.44 F1 score, 76.12 IoU score). This consistent superiority in $\text{F1}$ and $\text{IoU}$ scores across diverse benchmarks highlights UniChange's effectiveness and broad applicability.

\begin{table}[t]
\small
\setlength{\tabcolsep}{5.5pt}
\renewcommand\arraystretch{1.05}
\centering
\caption{Comparison with existing SOTA methods on SECOND. * denotes re-implementation under identical settings.}
\label{tab:scd}
\begin{tabular}{c|ccccc}
\toprule
Methods                                 & IoU   & $\text{mIoU}$ & $\text{F}_{scd}$  & $\text{F}_{bcd}$  & SeK   \\ \midrule
HRSCD.Str4   \cite{daudt2019multitask}  & 53.34     & 69.44   &   -   & 69.57     & 15.97 \\
ChangeMask   \cite{zheng2022changemask} & 54.23 & -       & -     & 70.32 & 17.89 \\
SAAN   \cite{guo2024saan}               & 53.49 & -       & -     & -     & 18.03 \\
DESNet   \cite{wang2023difference}      & -     & 70.73   & 58.75 & -     & 17.97 \\
STSP-Net   \cite{he2023spatial}         & -     & 72.03   & 60.77 & 72.05 & 20.91 \\
DFINet   \cite{wang2024dual}            & -     & 72.61   & -     & -     & 20.12 \\
MLFA-Net \cite{ding2024mlfa}            & 56.33 & 72.45   & -     & 72.06 & 20.11 \\
TextSCD   \cite{huang2025textscd}       & -     & 72.38   & 61.90 & -     & 21.66 \\
GLAI-Net   \cite{ding2025glai}          & 56.21 & -       & -     & 71.97 & 20.63 \\
Bi-SRNet*   \cite{ding2022bi}             & 56.89          & 72.25          & 62.12     & 72.52     & 21.69          \\
SCDNet*   \cite{peng2021scdnet}        & 55.51     & 71.63   & 62.40     & 71.39     & 21.43 \\
MTSCD-Net*   \cite{cui2023mtscd}         & 54.10 & 68.99   & 58.31 & 70.21 & 18.83 \\
SCD-SAM*   \cite{mei2024scd}             & 56.82 & 71.92   & 60.71 & 72.46 & 20.60 \\
HGINet*   \cite{long2024semantic}        & 56.13 & 71.73   & 62.88 & 71.90 & 21.83 \\
MambaSCD*   \cite{chen2024changemamba} & 57.24          & 72.73          & 54.17          & 72.81          & 21.31          \\ \rowcolor{blue!5}
\textbf{UniChange}                    & \textbf{57.62} & \textbf{72.85} & \textbf{63.50} & \textbf{73.12} & \textbf{23.02} \\ \bottomrule
\end{tabular}
\end{table}

\textbf{SCD Results. }As shown in \cref{tab:scd}, UniChange achieves the highest IoU score, mIoU score and $\text{F}_{bcd}$ score, confirming its superior capability in accurately delineating the overall regions of change and non-change. Furthermore, the model's SOTA performance on $\text{F}_{scd}$ and SeK metrics verifies that UniChange's semantic change perception capability is equally outstanding. This dual excellence across both binary and semantic accuracy capabilities exemplifies the unique, robust unification achieved by our MLLM-based framework, setting it apart from prior specialised models. For instance, MambaSCD, while achieving a strong binary detection result ($\text{mIoU}$ 72.73, $\text{F}_{bcd}$ 72.81), exhibits a substantial drop in semantic accuracy ($\text{F}_{scd}$ 54.17). Similarly, HGINet, despite attaining a competitive semantic score ($\text{F}_{scd}$ 62.88, the second highest), shows weaker binary detection performance ($\text{mIoU}$ 71.73, $\text{F}_{bcd}$ 71.90). This comprehensive superiority across all metrics confirms that UniChange's architecture is highly effective for handling the challenging task of multi-class change detection.

\subsection{Ablation Study}
In this section, we aim to validate the efficacy of the proposed components and strategies by conducting a series of ablation experiments. Unless otherwise specified, we employ LLaVA-7B-v1-1 as the base MLMM and RSBuilding-ViT-L as our vision backbone. Furthermore, semantic supervision is applied to the dual-temporal remote sensing images within the WHU-CD dataset. The lora rank is set to 8, and lora alpha is consistently set to twice the lora rank.

\textbf{Effects of Dual-Temporal Images Semantic Supervision.} The $\text{WHU-CD}$ dataset provides comprehensive annotations, including a change mask and individual semantic segmentation masks for both dual-temporal images ($\text{T}_1$ and $\text{T}_2$). To validate the effects of this dual-temporal semantic supervision, we conduct an ablation study categorizing experiments into four groups. As shown in \cref{abs:ss}, the optimal performance is realized by the final group, which applies semantic supervision to both temporal images ($\text{T}_1$ and $\text{T}_2$). The intermediate configurations (supervising only $\text{T}_1$ or $\text{T}_2$) achieve progressive performance improvements over the baseline (lacking semantic supervision), but are outperformed by the fully supervised setting. This monotonic trend confirms the synergistic effect of dual-temporal semantic supervision in enhancing feature discrimination, thereby affirming its efficacy within our unified framework.

\begin{table}[t]
\small
\setlength{\tabcolsep}{7pt}
\renewcommand\arraystretch{1.05}
\caption{Effects of dual-temporal images semantic supervision.}
\label{abs:ss}
\centering
\begin{tabular}{cc|cccc}
\toprule
\multicolumn{2}{c|}{Semantic Supervision} & \multicolumn{4}{c}{WHU-CD} \\ \midrule
$\phantom{100}\text{T}_1$ & $\text{T}_2$                 & P    & R    & F1   & IoU   \\ \midrule
\phantom{100}-                  & -                   & 95.38    & 93.51    & 94.42    & 89.46     \\
$\phantom{100}\checkmark$                   & -                & 95.77 & 93.45   & 94.59    &89.74           \\
\phantom{100}-                   & $\checkmark$                  & 95.16   & \textbf{94.39}    & 94.77    & 90.06     \\ \rowcolor{blue!5}
$\phantom{100}\checkmark$                   & $\checkmark$                  & \textbf{95.83}    & 94.11    & \textbf{94.96}    & \textbf{90.41}    \\ \bottomrule
\end{tabular}
\end{table}

\begin{figure*}[t]
  \centering
   \includegraphics[width=0.99\linewidth]{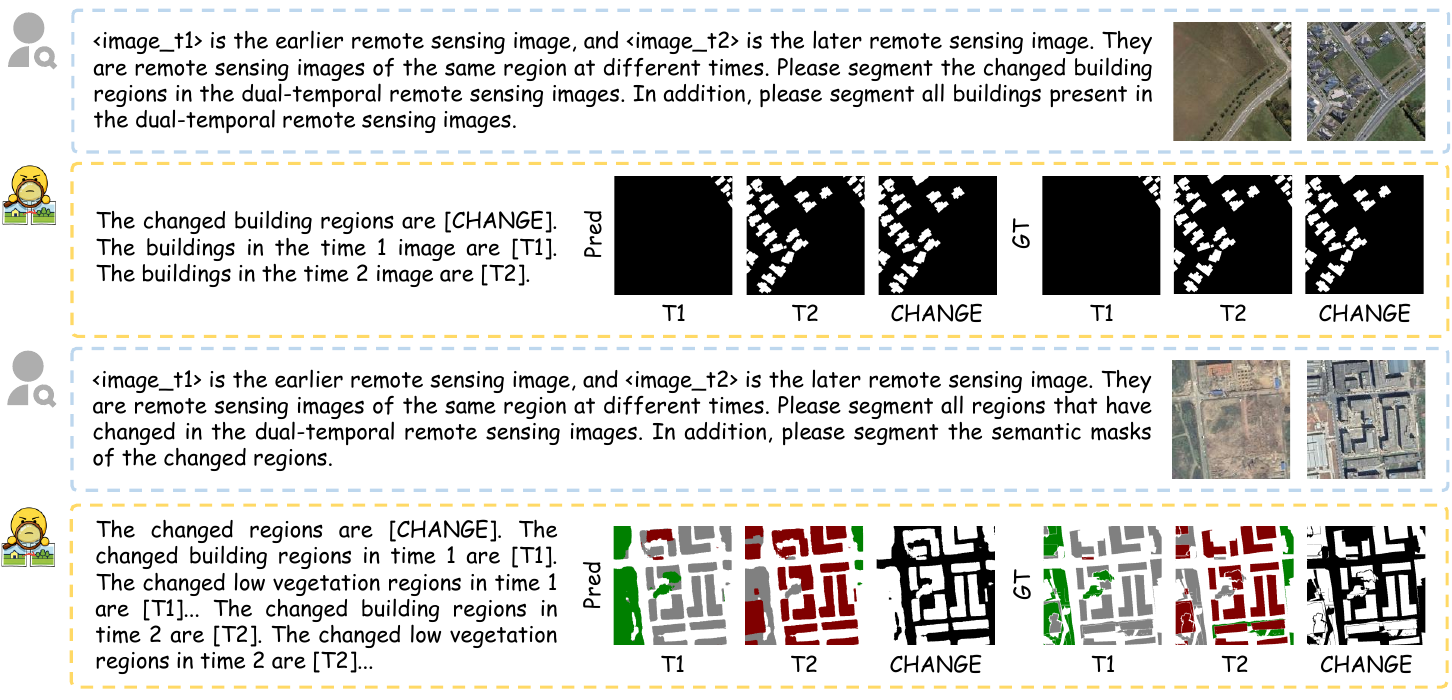}
   \caption{Visualisation results from UniChange. The images for the first question are sourced from the binary change detection dataset WHU-CD. The images for the second question are sourced from the semantic change detection dataset SECOND.}
   \label{fig:vis_results}
\end{figure*}

\textbf{Effects of Different Vision Backbones.} To validate the effects of different vision backbones, we conduct experiments using three backbones ($\text{SAM}$, $\text{SAM2}$, and $\text{RSBuilding-ViT-L}$) under frozen and fine-tuned ($\text{ft}$) configurations. Note that $\text{SAM}$ and $\text{RSBuilding}$ backbones share the same architecture, differing only in pre-trained weights. Results in \cref{abs:backbone} show two trends: First, the fine-tuned group consistently outperforms the frozen group. Second, $\text{RSBuilding-ViT-L}$ delivers best results in both settings.

\begin{table}[t]
\centering
\small
\setlength{\tabcolsep}{6.5pt}
\renewcommand\arraystretch{1.05}
\caption{Effects of different vision backbones.} 
\label{abs:backbone}
\begin{tabular}{c|cc|cc}
\toprule
\multirow{2}{*}{Vision Backbone} & \multicolumn{2}{c|}{SECOND} & \multicolumn{2}{c}{S2Looking} \\ \cmidrule(l){2-5} 

                           & SeK         & mIoU & F1 & IoU \\ \midrule
SAM \cite{kirillov2023segment}                        & 14.63         & 67.78                         & 59.51  & 42.36   \\
SAM2 \cite{ravi2024sam}                      & 13.84          & 67.48                        & 59.13   & 41.97   \\
RSBuilding-ViT-L \cite{wang2024rsbuilding}           & 19.30          & 71.11                 & 66.76  & 50.10   \\ \midrule
SAM(ft)              & 22.50 & 72.53                        & 66.87  & 50.23   \\
SAM2(ft)             & 22.82          & 72.63                        & 67.42  & 50.85   \\ \rowcolor{blue!5}
RSBuilding-ViT-L(ft) & \textbf{23.02}     & \textbf{72.85}    & \textbf{69.32}  & \textbf{53.04}   \\ \bottomrule
\end{tabular}
\end{table}

\textbf{Effects of LoRA Ranks. }To validate the effects of different LoRA ranks on model performance, we conduct experiments with LoRA ranks of 4, 8, 16, and 32, with the results detailed in \cref{abs:lr}. As shown in the table, the model's performance varies with different LoRA ranks, achieving optimal results when the LoRA rank is set to 8.
\begin{table}[t]
\centering
\small
\setlength{\tabcolsep}{11pt}
\renewcommand\arraystretch{1.05}
\caption{Effects of different LoRA ranks.}
\label{abs:lr}
\begin{tabular}{c|cc|cc}
\toprule
\multirow{2}{*}{LoRA Rank} & \multicolumn{2}{c|}{WHU-CD} & \multicolumn{2}{c}{S2Looking} \\ \cmidrule(l){2-5} 
                           & F1         & \multicolumn{1}{c|}{IoU} & F1 & IoU \\ \midrule
4                        & 94.68          & 89.89            & 68.53  & 52.13   \\ \rowcolor{blue!5}
8                       & \textbf{94.96}          & \textbf{90.41}                        & \textbf{69.32}  & \textbf{53.04}   \\
16           &94.82          & 90.15                        & 68.42  & 52.00   \\
32 & 94.81         & 90.13                        & 68.16  & 51.70   \\ \bottomrule
\end{tabular}
\end{table}

\textbf{Effects of Joint Training Datasets.} We design a progressive experimental setup to validate the effects of joint training with multiple datasets, with results presented in \cref{abs:joint}. In this study, A, B, C, and D represent the $\text{WHU-CD}$, $\text{S2Looking}$, $\text{LEVIR-CD+}$, and $\text{SECOND}$ datasets, respectively. We progressively increase the number of datasets in the training mixture, starting with A, then A+B, A+B+C, and finally A+B+C+D. Results show a clear upward trend in overall performance as more datasets are added. This indicates that joint training allows the model to learn more general and robust representations. The best results for all datasets are achieved when the model is trained jointly on all four datasets.

\begin{table}[t]
\centering
\small
\setlength{\tabcolsep}{5pt}
\renewcommand\arraystretch{1.05}
\caption{Effects of joint training datasets.}
\label{abs:joint}
\begin{tabular}{cccc|c|c|c|cc}
\toprule
\multicolumn{4}{c|}{Training Dataset} & A   & B   & C   & \multicolumn{2}{c}{D} \\ \midrule
A       & B       & C       & D       & IoU & IoU & IoU & mIoU      & SeK      \\ \midrule
$\checkmark$       &  -       & -        & -        &89.68     & -    & -    & -          & -          \\
$\checkmark$       & $\checkmark$       &  -       &-         & 89.77    & 52.57    &-     & -          & -          \\
$\checkmark$       & $\checkmark$      & $\checkmark$       & -        &90.16     &52.51     &78.43     & -          &  -         \\ \rowcolor{blue!5}
$\checkmark$       & $\checkmark$       & $\checkmark$       & $\checkmark$       & \textbf{90.41}    &\textbf{53.04}     &  \textbf{78.87}   &  \textbf{72.85}         &  \textbf{23.02}         \\ \bottomrule
\end{tabular}
\end{table}

\subsection{Visualisation Results}
As illustrated in \cref{fig:vis_results}, we present the visualisation results for UniChange. It can be observed that UniChange demonstrates commendable performance across both BCD and SCD tasks. Further detailed visualisation results are provided in \cref{appendix:vis} of the supplementary materials.
\section{Conclusion}
\label{sec:conclusion}
In this paper, we introduce UniChange, the first unified change detection framework built upon MLLM. This framework addresses the fundamental challenges of task fragmentation and dataset incompatibility that constrain conventional change detection models. 
Our approach reframes both BCD and SCD tasks as ``embedding as change''. By querying semantics or changes through the embeddings of three special tokens ([T1], [T2], and [CHANGE]), UniChange eliminates the need for predefined classification headers. This design uniquely allows the model to acquire comprehensive knowledge by jointly training on diverse, multi-source datasets, even when their semantic labels conflict. Our approach achieves state-of-the-art performance on four public benchmarks (WHU-CD, S2Looking, LEVIR-CD+, and SECOND), demonstrating its superiority.
{
    \small
    \bibliographystyle{ieeenat_fullname}
    \bibliography{main}
}
\clearpage
\appendix
\twocolumn[
  \begin{center}
    \Large \textbf{UniChange:~Unifying~Change~Detection~with~Multimodal~Large~Language~Model} \\ 
    \vspace{1em}
    \textbf{Supplementary Material} \\
  \end{center}
  \vspace{2em}
]
\section*{Overview}
This supplementary material provides additional details to support the main manuscript. The document is structured as follows: \cref{appendix:eva_met} begins with a detailed breakdown of the evaluation metrics, covering the specific formulas used for both BCD and SCD tasks. Following this, \cref{appendix:vis} offers extensive qualitative visual comparisons, benchmarking our model against other state-of-the-art methods on several key datasets. Finally, \cref{appendix:datasets_datails} presents a comprehensive description of the datasets utilised in our experiments, including their technical specifications and key characteristics.

\section{Evaluation Metrics}
\label{appendix:eva_met}
\subsection{BCD Metrics}
To evaluate the performance of UniChange on the binary change detection (BCD) task, we employ four standard pixel-level metrics: Precision (P), Recall (R), F1 Score (F1), and Intersection over Union (IoU). Precision measures the proportion of correctly predicted change pixels among all pixels classified as change. Recall indicates the proportion of true positive pixels among all truly positive pixels in the ground truth. The F1 Score is the harmonic mean of Precision and Recall, providing a single metric that balances these two measures. Finally, IoU measures the overlap between predicted and ground-truth positive regions, serving as a robust measure of segmentation quality.

The metrics are individually defined as follows, where TP, TN, FP, and FN represent the number of true positive pixels, true negative pixels, false positive pixels, and false negative pixels, respectively:
\begin{equation}
\begin{aligned}
    \text{P} &= \text{TP / (TP + FP)}, \\
    \text{R} &= \text{TP / (TP + FN)}, \\
    \text{F1} &= \text{2} \times \text{P} \times \text{R} / (\text{P} + \text{R}), \\
    \text{IoU} &= \text{TP / (TP + FP + FN)}.
\end{aligned}
\end{equation}

\subsection{SCD Metrics}
The assessment of semantic change detection (SCD) model performance is executed using a collection of specialised metrics, all derived from the confusion matrix $Q = \{q_{ij}\}$, where $q_{ij}$ records the count of pixels classified as class $i$ with a ground truth label of $j$.

The mean Intersection over Union for SCD ($\text{mIoU}$) is utilised to evaluate overall segmentation quality, established as the arithmetic mean of the Intersection over Union for the regions without change ($\text{IoU}_{nc}$) and all changing regions ($\text{IoU}_{c}$):
\begin{equation}
\text{mIoU} = (\text{IoU}_{nc} + \text{IoU}_{c}) / 2.
\end{equation}

$\text{IoU}_{nc}$ measures the overlap between the predicted unchanged regions and the ground-truth unchanged regions:
\begin{equation}
\text{IoU}_{nc} = q_{00} / (\sum_{i=0}^{N}q_{i0} + \sum_{j=0}^{N}q_{0j} - q_{00}).
\end{equation}

Conversely, $\text{IoU}_{c}$ measures the overall segmentation quality of all change regions, treating all distinct semantic change categories as a single change class:
\begin{equation}
\text{IoU}_{c} = \sum_{i=1}^{N}\sum_{j=1}^{N}q_{ij} / (\sum_{i=0}^{N}\sum_{j=0}^{N}q_{ij} - q_{00}). 
\end{equation}

The Separation kappa coefficient (SeK) provides a valuable measure of semantic discrimination amidst class imbalance, particularly designed to diminish the influence of the prevalent unchanged class. SeK is computed from the confusion matrix $\hat{Q} = \{\hat{q_{ij}}\}$, where $\hat{q}_{ij} = q_{ij}$, but $\hat{q}_{00}=0$, and is calculated as
\begin{equation} 
\begin{gathered}
\rho = \sum_{i=0}^{N} \hat{q}_{ii} \bigg/ \sum_{i=0}^{N} \sum_{j=0}^{N} \hat{q}_{ij}, \\
\eta = \sum_{i=0}^{N} \left( \sum_{j=0}^{N} \hat{q}_{ij} \times \sum_{j=0}^{N} \hat{q}_{ji} \right) \bigg/ \left( \sum_{i=0}^{N} \sum_{j=0}^{N} \hat{q}_{ij} \right)^2, \\
SeK = e^{IoU_{c}-1} \cdot (\rho - \eta) / (1 - \eta).
\end{gathered}
\end{equation}

The F1-Score for SCD ($\text{F}_{scd}$) offers a focused quantification of semantic transition accuracy within change regions. This metric is derived from the precision ($\text{P}_{scd}$) and recall ($\text{R}_{scd}$) over the pixels determined to have changed. 

The SCD precision $\text{P}_{scd}$ and recall $\text{R}_{scd}$ are defined as
\begin{equation} 
\begin{gathered}
P_{scd} = \sum_{i=1}^{N}q_{ii} / \sum_{i=1}^{N}\sum_{j=0}^{N}q_{ij},
\\
R_{scd} = \sum_{i=1}^{N}q_{ii} / \sum_{i=0}^{N}\sum_{j=1}^{N}q_{ij}.
\end{gathered}
\end{equation}

The final $\text{F}_{scd}$ is the harmonic mean of these two components:
\begin{equation}
F_{scd} = 2 \times P_{scd} \times R_{scd} / (P_{scd} + R_{scd}). 
\end{equation}

\twocolumn[
\centering 
\captionof{table}{Detailed information about change detection datasets used for experiments.}
\label{tab:dataset_details}
\begin{tabular}{c|ccccc} 
\toprule
Dataset & Resolution & Image Size & Image Number & Evaluation Task & Class\\ \midrule
WHU-CD & 0.075m & $\text{32507} \times \text{15354}$ & 1 & BCD & Building\\
S2Looking  & 0.5-0.8m & $\text{1024} \times \text{1024}$ & 5000 & BCD & Building\\
LEVIR-CD+  & 0.5m & $\text{1024} \times \text{1024}$ & 985 & BCD & Building\\ \midrule
\multirow{2}{*}{SECOND} & \multirow{2}{*}{0.5-3m} & \multirow{2}{*}{$\text{512} \times \text{512}$} & \multirow{2}{*}{4662} & \multirow{2}{*}{SCD} & Building, Low Vegetation, Tree, \\
 & & & & & Water, Playground, Bare Ground\\ 
\bottomrule
\end{tabular}
\vspace{20pt}
]

The performance of UniChange on the SCD task is comprehensively evaluated using BCD metrics: IoU and F1 ($\text{F}_{bcd}$) and SCD metrics: mIoU, $\text{F}_{scd}$ and SeK.

\section{Visual Comparisons}
\label{appendix:vis}
To qualitatively evaluate the performance of our proposed UniChange, we provide comprehensive visual comparisons against other state-of-the-art (SOTA) methods on both binary and semantic change detection tasks.

\begin{figure*}[t]
  \centering
   \includegraphics[width=0.99\linewidth]{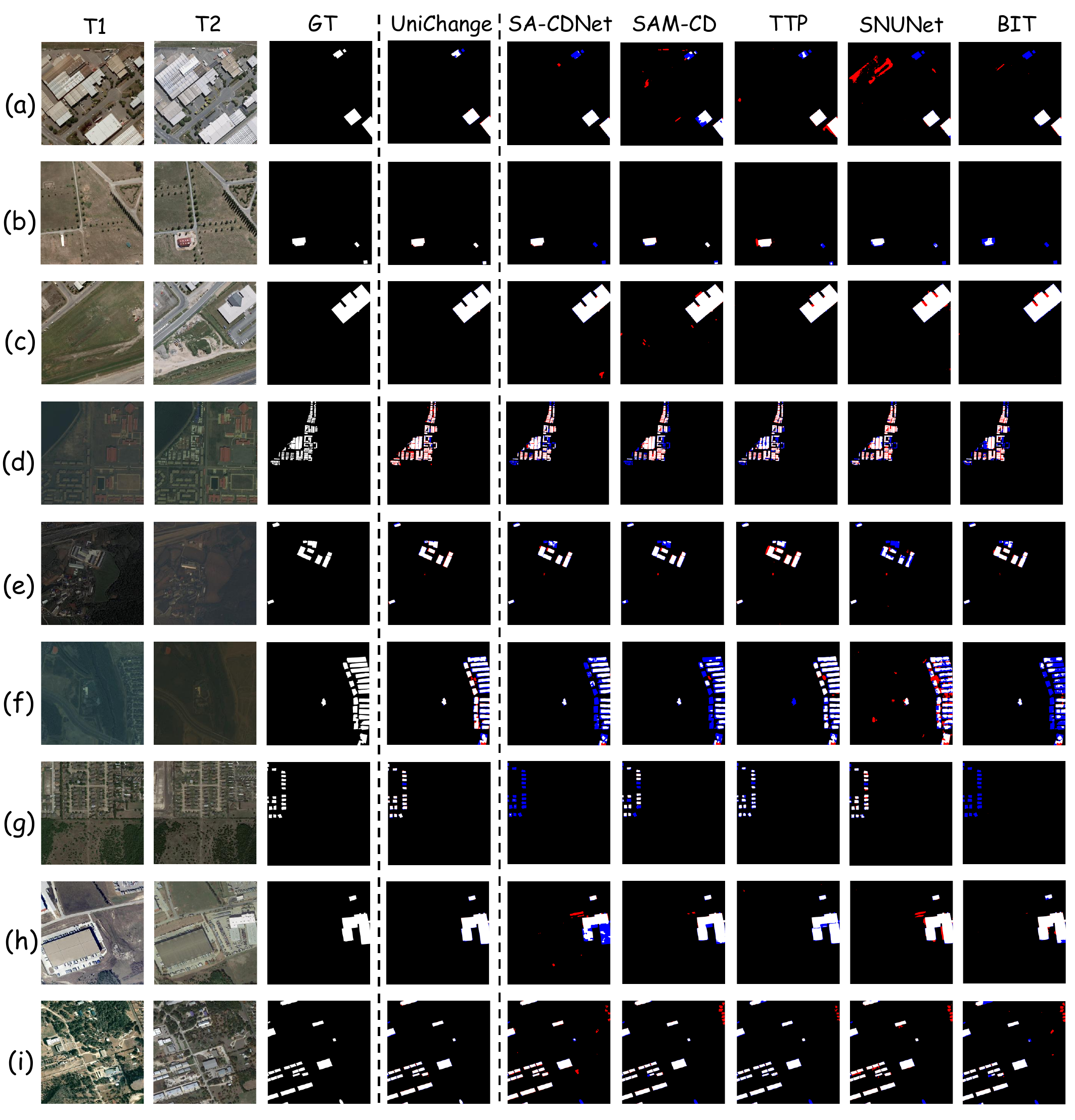}
   \caption{Visual comparisons of the UniChange with other state-of-the-art methods for binary change detection. {\color{red}\textbf{Red}} means false positives (FP), while {\color{blue}\textbf{Blue}} denotes false negatives (FN). Samples (a) (b) (c) are from the WHU-CD dataset, (d) (e) (f) are from the S2Looking dataset, and (g) (h) (i) are from the LEVIR-CD+ dataset.}
   \label{fig:appendix_vis_bcd}
\end{figure*}

\cref{fig:appendix_vis_bcd} illustrates the visual results for binary change detection (BCD). The comparisons on the WHU-CD, S2Looking, and LEVIR-CD+ datasets all show that UniChange performs better than competing methods. Its generated change masks are more complete and have fewer false positives (FP) and false negatives (FN).

\begin{figure*}[t]
  \centering
   \includegraphics[width=0.99\linewidth]{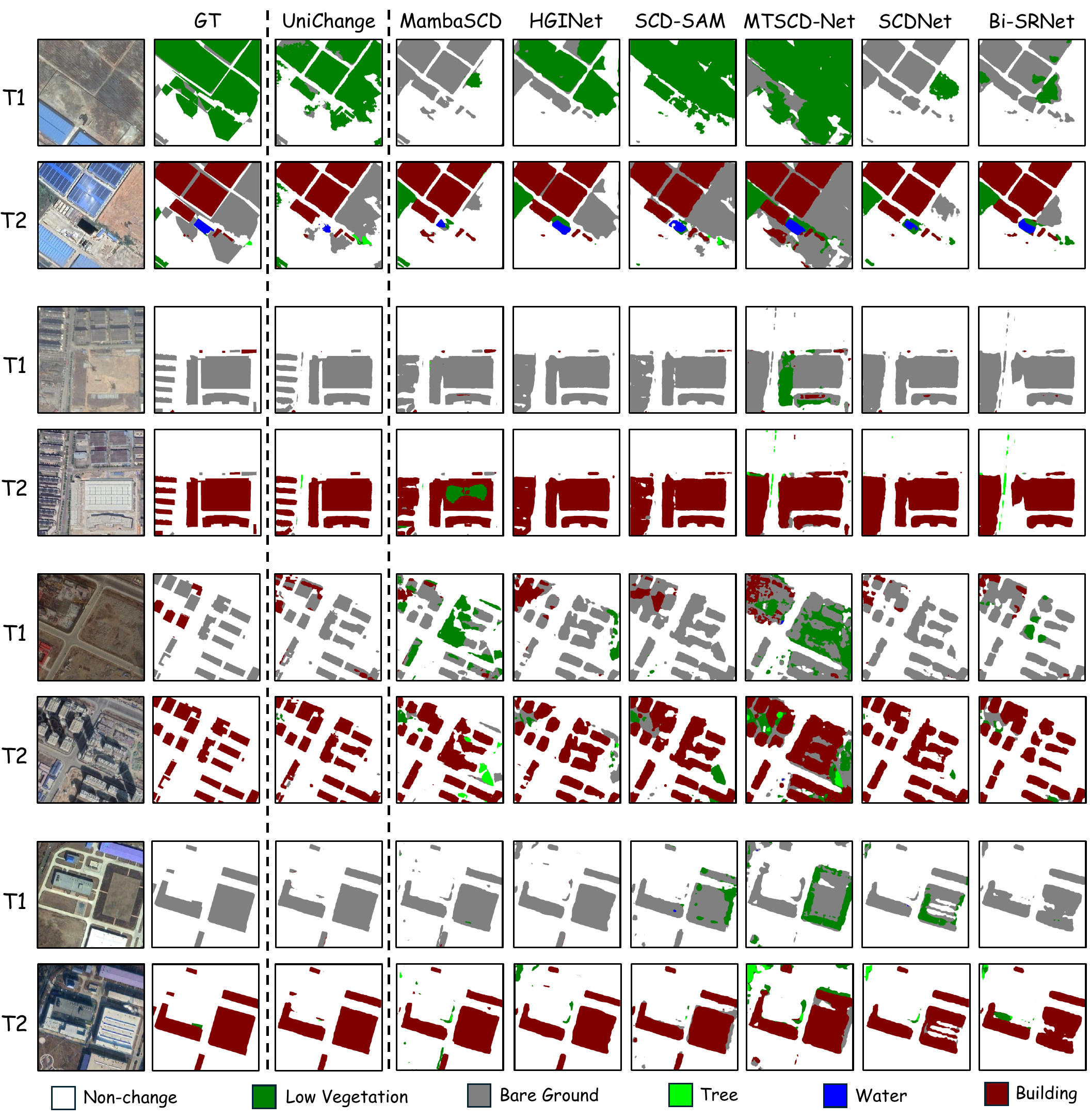}
   \caption{Visual comparisons of the UniChange with other state-of-the-art methods for semantic change detection. The colour legend is as follows: \textbf{White} represents Non-change, {\color{LegendDarkGreen}\textbf{dark green}} represents Low Vegetation, {\color{LegendGray}\textbf{grey}} represents Bare Ground, {\color{LegendBrightGreen}\textbf{bright green}} represents Tree, {\color{LegendBlue}\textbf{blue}} represents Water, and {\color{LegendDarkRed}\textbf{dark red}} represents Building.}
   \label{fig:appendix_vis_second}
\end{figure*}

Furthermore, \cref{fig:appendix_vis_second} presents the qualitative results for semantic change detection (SCD). As shown in the figure, UniChange not only accurately locates the changed regions but also exhibits a strong capability in correctly identifying the specific semantic categories of the changes. These visual results underscore the superior performance and generalisation ability of our unified model.

\section{Dataset Details}
\label{appendix:datasets_datails}
To comprehensively evaluate our model, we utilise four diverse, publicly available remote sensing datasets, covering both binary change detection (BCD) and semantic change detection (SCD) tasks. The detailed specifications of these datasets are summarised in \cref{tab:dataset_details}.

For the BCD task, we utilise three datasets, all of which focus on identifying building changes. The WHU-CD dataset provides a single, massive-scale image pair ($32507 \times 15354$ pixels), captured at an ultra-high 0.075m resolution. It is particularly notable for its coverage, capturing a 20.5 km$^2$ area in Christchurch, New Zealand. The dual-temporal images, captured in 2012 and 2016, document the region's reconstruction following a major earthquake. This provides an excellent case for studying large-scale urban recovery. In addition, we use S2Looking, a large-scale dataset consisting of 5000 dual-temporal image pairs, which are $1024 \times 1024$ pixels in size and have a resolution of 0.5-0.8m. Its defining characteristic is the use of satellite side-looking images captured at various off-nadir angles, a sharp contrast to typical near-nadir (or top-down) imagery. This dataset focusses on globally distributed rural areas. It challenges models with large illumination variances and complex rural scenes. It also features geometric distortions from the oblique angles. Finally, we utilise LEVIR-CD+, an expanded version of the LEVIR-CD dataset. It contains 985 image pairs of 0.5m resolution Google Earth imagery. Each image pair is $1024 \times 1024$ pixels. In contrast to S2Looking, LEVIR-CD+ features near-nadir images and focusses on building changes in urban and suburban environments, primarily covering 20 different regions in Texas, USA.

For the more complex SCD task, we use the SECOND dataset. This is a large-scale benchmark containing 4,662 pairs of aerial images, each $512 \times 512$ pixels in size. The images come from several platforms and sensors. They cover major cities in China, including Hangzhou, Chengdu, and Shanghai. Unlike the BCD datasets, SECOND requires the model to identify semantic transitions among six distinct land-cover classes: building, low vegetation, tree, water, playground, and bare ground. A defining feature of this dataset is its annotation method; it provides land-cover map pairs and nonchange masks. This structure is specifically designed to challenge models. It tests their ability to detect changes that occur between the same land-cover class. For example, a model must find where an old playground is removed and a new one is built in its place. This is a critical capability that many other datasets cannot evaluate.

\clearpage


\end{document}